\documentclass{article}
\usepackage{iclr2022_conference,times}

\usepackage{amsmath}
\usepackage{graphicx}
\usepackage{subcaption}
\usepackage{hyperref}
\usepackage{cleveref}
\usepackage{rotating}

\DeclareMathOperator{\acc}{ACC}
\newcommand{\msa}{\text{MSA}}

\title{Machine Learning State-of-the-Art with Uncertainties}

\author{Peter Steinbach, Felicita Gernhardt, Mahnoor Tanveer, Steve Schmerler, Sebastian Starke \thanks{ML Evaluation Standards Workshop at ICLR 2022.} \\
Department of Computational Science\\
Core Facility for Information Services and Computing\\
Helmholtz-Zentrum Dresden-Rossendorf \\
\texttt{\{p.steinbach, f.gernhardt,  m.tanveer, s.schmerler, s.starke\}@hzdr.de}
}

\iclrfinalcopy % Uncomment for camera-ready version, but NOT for submission.
\begin{document}

\maketitle

\begin{abstract}
With the availability of data, hardware, software ecosystem and relevant skill sets, the machine learning community is undergoing a rapid development with new architectures and approaches appearing at high frequency every year. In this article, we conduct an exemplary image classification study in order to demonstrate how confidence intervals around accuracy measurements can greatly enhance the communication of research results as well as impact the reviewing process. In addition, we explore the hallmarks and limitations of this approximation. We discuss the relevance of this approach reflecting on a spotlight publication of ICLR22. A reproducible workflow\footnote{Our workflow is available on \href{https://github.com/psteinb/sota_on_uncertainties/}{github.com/psteinb/sota\_on\_uncertainties}.} is made available as an open-source adjoint to this publication. Based on our discussion, we make suggestions for improving the authoring and reviewing process of machine learning articles.
\end{abstract}

\section{Introduction}
\label{sec:intro}

The machine learning (ML) community has established a routine during the publication life cycle in which new approaches are first validated on benchmark datasets in experiments. During paper review, the viability of propositions is reasoned based on these experiments and compared to current state-of-the-art results. Model comparison is henceforth a common task during the review and authoring process.

With the availability of training data -- like ImageNet by \citet{ImageNet}, qoca by \citet{DBLP:journals/corr/abs-1808-07042}, bbbc by \citet{BBC004:4567415} -- in conjunction with available hardware and human software skills, established conferences are flooded by submissions. Studies like \citet{confsubs} report submission numbers above $1000$ every year for the major conferences in the field. This demand has called for more reviewing capacity provided by the conferences in turn. Given the acceptance rates of submitted articles, new approaches for common ML tasks are available every year.

In parallel, the spread of ML techniques is still ongoing in all domains of society, industry and academia. In these application domains, new challenges of dataset availability and heterogeneity await any data scientist or ML practitioner who wants to exploit ML in her or his field. These practitioners need to decide where to put their time. Having publications being released with publicly available code reduces the barrier to try out new architectures. Model comparison is henceforth a common task for practitioners as well.

In the following discussion, we offer our suggestions for evaluation of ML architectures to eventually augment the process of model comparison:

\begin{itemize}
\item We suggest a minimal standard to report ML experiments (based on an analysis available as an open-source repository in \citet{ourrepo}).
\item When evaluating claims based on global performance metrics, we suggest uncertainty (or variance) and its approximation as a first-class citizen in the review process.

\end{itemize}

\section{Experiments}
\label{sec:experiments}

Rather than providing a meta-analysis of already published results and discussions, we provide a representative example analysis at first. We use this to put the discussion not only in the perspective of the reviewer but also in the perspective of the submitting authors. Given our motivation from \Cref{sec:intro}, we believe that both sides should be heard in this discussion.

We base our argumentation on the task of image classification as an example. We use the popular \texttt{timm} library \citep{timm} to train and run image classification models. In order to allow reproducible results, all steps in our analysis are encoded in an automated snakemake \citep{smk} workflow available as open-source \citep{ourrepo}. All details on our experiments are given in the Appendix \Cref{ssec:exp_details}. Our reproducible workflow is illustrated in Appendix \Cref{ssec:smk_fold10}.

As accuracy is the predominant metric in use in image classification published at ML conferences, we use it here as a representative example and report it after training for 80 epochs.

To illustrate our point we set out to compare the performance of three different architectures. We report their measured accuracy values (computed on the holdout validation set) side by side to learn which architecture offers the highest performance as is often done in publications. To obtain a more comprehensive view, we take the discussion in section 1.7 of \citet{raschka2018model} and compute a confidence interval around the accuracy point estimate under the \textit{normal approximation}. The core idea behind this approach considers accuracy as a random variable which complies to a binomial distribution over the holdout folds available. The derivation of \citet{raschka2018model} concludes the estimated confidence interval to be

\begin{equation}
\label{eq:accconf}
    \hat{\sigma} = z\,\sqrt{\frac{1}{n_{holdout}}\,\acc_{holdout}\,(1 - \acc_{holdout}) }
\end{equation}

In \cref{eq:accconf}, $z$ denotes the $1 - \frac{\alpha}{2}$ error quantile of a standard normal distribution with $\alpha$ being the error quantile itself. We chose a one-sigma confidence interval of $68.1 \%$ and used $z = 1$ (a $95 \%$ confidence interval requires $z = 1.96$). $n_{holdout}$ refers to the number of samples in the holdout validation set and $\acc_{holdout}$ to the accuracy calculated therein.

\begin{figure}[h]
    \centering
    \begin{subfigure}[b]{0.48\textwidth}
        \includegraphics[width=\textwidth]{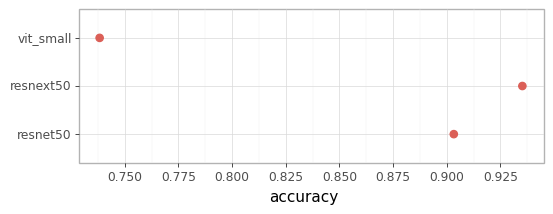}
        \caption{Point estimates}
        \label{fig:acc_point}
    \end{subfigure}
    ~ %add desired spacing between images, e. g. ~, \quad, \qquad, \hfill etc.
    \begin{subfigure}[b]{0.48\textwidth}
        \includegraphics[width=\textwidth]{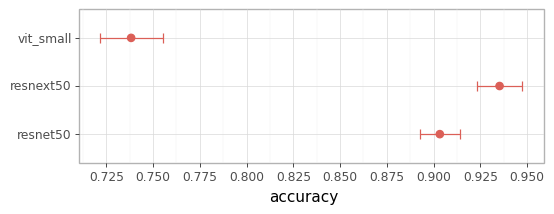}
        \caption{Point estimates and confidence intervals}
        \label{fig:acc_err}
    \end{subfigure}
    \caption{Accuracy estimates on 10-class image classification for three different ML architectures. \Cref{fig:acc_point} shows point estimates only. \Cref{fig:acc_err} shows point estimates and confidence intervals computed from sample standard deviations of accuracies in 20-fold cross-validation.}
    \label{fig:accuracies}
\end{figure}

From \cref{fig:accuracies} we learn that the confidence estimates of both ResNe(X)T architectures are very close to each other. One-sigma confidence intervals of both models almost contact each other. Given the fact that these are one-sigma confidence intervals for accuracy, the performance difference between the two ResNe(X)T models can and should never be reported as significant. In contrast, the Vision Transformer implementation offers distinctly inferior accuracy. Note, a rigorous conclusion whether reported accuracies are significantly different from each other requires mathematical tools such as McNemar's test \citep{mcnemar1947note} or similar which are beyond the reach of this article.

To empirically confirm that the approximation of uncertainties stated above does hold, we created $20$ stratified folds of the \texttt{imagenette} dataset and repeated training for $80$ epochs again on all three model architectures as suggested in \citet{raschka2018model}. Each fold offered a hold-out dataset of $670$ images of equal proportion for each class. In addition, we performed the training with six different seeds on all available folds in order to include effects from initialization of the network weights as reported in \citet{hooker}. The results of this analysis are available in Appendix \Cref{sssec:confcompare}. The latter study provides us confidence that \Cref{eq:accconf} is a valid approximation to describe variability of obtained accuracy measurements in practice. At the same time, we highlight limitations of this approach.

Uncertainty estimates enhance display of ML results and help judge the degree of progress offered. The approximation technique illustrated in this section provides an author with a time and hardware runtime effective technique to quantify uncertainty estimates of ML metrics perceived as random variables. Finally, having uncertainties displayed provides an intuition to readers and reviewers on how robust the offered findings are compared to the established state-of-the-art in the field.

\section{Application}
\label{sec:application}

In order to support our argumentation, we would like to reflect on a peer-reviewed paper published in ICLR22. It was awarded as a spotlight paper. The authors have provided their code on a public platform \citep{howvitswork_code} and all review comments have been made available \citep{howvitswork_review}. We like to point out that this is a singular piece of evidence and hence should not serve to represent the entire ensemble of ML articles submitted to any ML conference.

In the paper ``How Do Vision Transformers Work?" \citep{howvitswork}, the main discussion revolves around the observation that vision transformer networks (exploiting multi-head self-attention, MSA) for image classification appear to smooth the loss landscape. The article proposes a new network design based on this (AlterNet) and claims that AlterNet outperforms CNNs.

\begin{figure}[h]
    \centering
    \begin{subfigure}[b]{0.48\textwidth}
        \includegraphics[width=\textwidth]{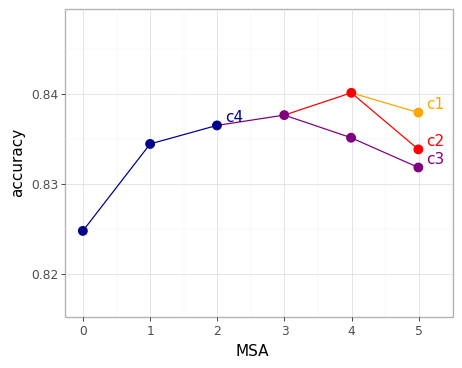}
        \caption{No uncertainties}
        \label{fig:refig12a_noerrors}
    \end{subfigure}
    ~ %add desired spacing between images, e. g. ~, \quad, \qquad, \hfill etc.
    \begin{subfigure}[b]{0.48\textwidth}
        \includegraphics[width=\textwidth]{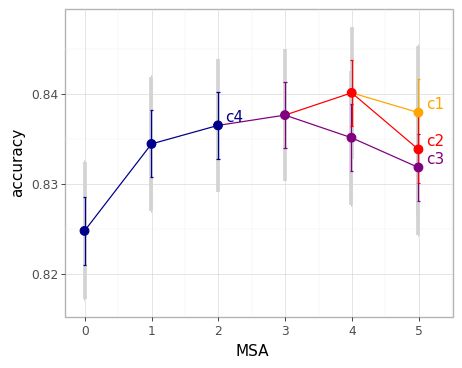}
        \caption{Including uncertainties}
        \label{fig:refig12a_errors}
    \end{subfigure}
    \caption{Reproduction of figure 12a from \citet{howvitswork} (left). Augmentation of the same figure with estimated accuracy calculated using \cref{eq:accconf} using a one-sigma $68.2 \%$ (colored) and two-sigma $95 \%$ (grey) confidence interval (right). Data to reproduce these figures was obtained by using \citet{wpd} on the figures from the preprint PDF.
    }

    \label{fig:refig12a}
\end{figure}

\Cref{fig:refig12a} reproduces and augments Figure 12a from \citet{howvitswork}. In the original text, the article concludes:

\begin{quote}
    Figure 12a reports the accuracy of Alter-ResNet-50, which replaces the Conv blocks in ResNet-50 with local MSAs \dots As expected, MSAs in the last stage (c4) \textbf{significantly} improve the accuracy. Surprisingly, an MSA in 2nd stage (c2) improves the accuracy, while two or more MSAs in the 3rd stage (c3) reduce it. In conclusion, MSAs at the end of a stage play an important role in prediction.
\end{quote}

Augmenting the results with approximated confidence intervals uncovers and underlines the intrinsic uncertainty of the presented results in visual form. As these accuracies were obtained based on a single holdout set, the assumptions of the normal approximation of \cref{eq:accconf} hold.

Further, the performance gain using MSA elements (any entry with $\msa > 0$) versus a classical network design without them ($\msa=0$) is reduced when taking the uncertainties into account. It is surprising to note that none of the four published reviews commented on this shortcoming of the discussion in the paper. This may indicate that AlterNet may not offer distinct performance gains as suggested in the text above. When considering the $95 \%$ confidence intervals (grey shaded intervals in \cref{fig:refig12a_errors}), all confidence intervals for $\msa > 0$ overlap with the one for $\msa=0$. From our point of view, this has a considerable impact on the conclusions brought forward by this article.

\section{Conclusion}
\label{sec:conclusion}

Reporting known uncertainties on metrics obtained in ML experiments is an important but challenging task. Nonetheless it is indispensable for the evaluation of ML experiments. Uncertainties are essential especially as the progress of improvements in the field has slowed down in recent years \citep{Bouthillier}. The core challenge does not only reside in the computational expense required (see multi-fold, multi-seed experiments performed and summarized in \cref{fig:compare_noseed_approx}), but also in the statistical treatment of correlations (which we did not discuss in this article).

In this spirit, we showed that approximations to uncertainties can be applied to standard use cases often seen in the community. In this way, they provide an important tool for reviewers and readers alike to judge the presented results and relate their degree of progress. Uncertainty estimates help to establish trustworthiness in any author's results.

We emphasize that reporting uncertainties is only a first step towards rigorous model comparison using hypothesis testing based on McNemar's test \citep{mcnemar1947note} or similar approaches. While we believe that hypothesis testing based model comparison with high statistical power should be the norm, we acknowledge its practical implications. Hence, we strongly encourage the use of uncertainties and their approximations as an intermittent and essential step towards this goal in order to support reviewers and authors alike.

We suggest that conferences and journals invest more effort in on-boarding reviewers to be aware of these statistical tools. Moreover, journals and conferences should make sure that authors report the central parameters required for computing such approximations. In an extreme case, articles could also be desk rejected if they fall short of delivering parameters for quantifying uncertainties. In the best of all worlds, the community could suggest conferences and journals to make reporting uncertainties a requirement for submission. Along this line of thought, the community could define a single or tiered levels of standards with degrees of confidence such intervals should encode. With such measures, we hope that our community can promote uncertainty to a first class citizen in the evaluation of ML models and experiments.

Last but not least, we have demonstrated that such investigations are possible with today's software tools to an extent that offers a scalable software environment allowing authors to use cloud or HPC infrastructures (e.g. for model training). At the same time, this technology yields provenance information of results from the training data to the final plot of a publication. Publishing software alongside articles is a good first step towards reproducibility. We believe that the use of such workflow software should be encouraged in the submission process to ensure reproducibility and transparency at the same time.

\section*{Acknowledgments}

The authors gratefully acknowledge the computing time granted on the supercomputer JUSUF \citep{jusuf} at Forschungszentrum Jülich. This work is supported by the Helmholtz Association Initiative and Networking Fund under the Helmholtz AI platform grant.

\bibliography{main}
\bibliographystyle{iclr2022_conference}

\section{Appendix}
\label{sec:appendix}

\subsection{Experimental Details}
\label{ssec:exp_details}

As a training dataset, we use \texttt{imagenette} \citep{imagenette}, a subset of ImageNet \citep{ImageNet} which amounts to $n_{train} = 9469$ training images of $320\times 320$ pixels encoded as RGBA JPEG files available in 10 distinct object classes. The holdout validation set sums up to $n_{holdout}=3925$ images.

We use the popular \texttt{timm} library \citep{timm} to train and run three image classification models. In order to allow reproducible results, all steps in the analysis are encoded in an automated snakemake \citep{smk} workflow.

The three architectures of choice used in \cref{sec:experiments} are as follows:

\begin{itemize}
\item \texttt{restnet} is a ResNet architecture \citep{resnet}, \texttt{timm} model: \texttt{resnet50}
\item \texttt{resnext} is a ResNext architecture \citep{resnext}, \texttt{timm} model: \texttt{resnext50\_32x4d}
\item \texttt{vit\_small} is a Vision transformer \citep{vit}, \texttt{timm} model: \texttt{vit\_small\_patch32\_224}
\end{itemize}

All have been trained using the default parameters in \texttt{timm} version $0.5.4$ without using pretrained weights.

To empirically confirm that the approximation of uncertainties stated above does hold, we created $20$ stratified folds of the \texttt{imagenette} dataset and repeated training for $80$ epochs again on all three model architectures as suggested in \citet{raschka2018model}. Each fold offered a hold-out dataset of $670$ images of equal proportion for each class. In addition, we performed the training with six different seeds on all available folds.

All code to reproduce experiments is made available \citep{ourrepo}.

\subsection{Multi-Seed Multi-Fold Experiments}
\label{ssec:multiseed}

\subsubsection{Distribution of Accuracies}
\label{sssec:histos}

In this section, we provide more insights into the distribution of accuracy measurements across seeds and folds.

\begin{figure}[h]
    \centering
    \begin{subfigure}[b]{\textwidth}
        \includegraphics[width=\textwidth]{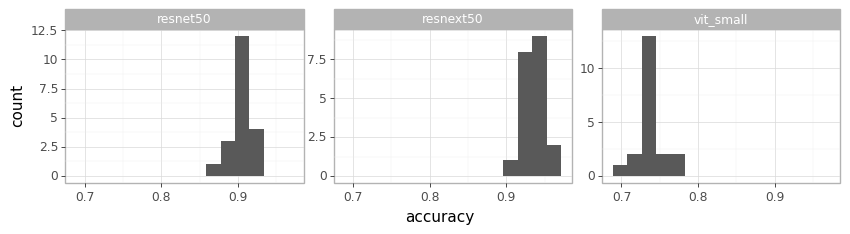}
        \caption{One seed}
        \label{fig:seed42}
    \end{subfigure}
    \newline
    \begin{subfigure}[b]{\textwidth}
        \includegraphics[width=\textwidth]{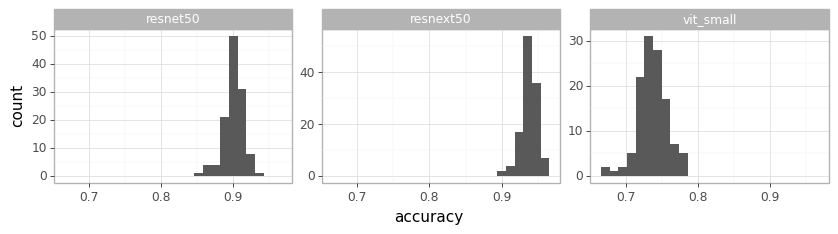}
        \caption{Six different seeds}
        \label{fig:allseeds}
    \end{subfigure}
    \caption{Histogram of accuracy estimates for 10-class image classification for three different ML architectures across 20 folds and 6 seeds. \Cref{fig:seed42} shows results for one seed only (15 bins due to low statistics). \Cref{fig:allseeds} shows results across all 6 seeds (25 bins).}

    \label{fig:acc_histos}
\end{figure}

As suggested in \citet{raschka2018model}, the histograms obtained in \cref{fig:acc_histos} reveal an asymmetric structure while being centered at the reported mean values. Especially \cref{fig:allseeds} reveals interesting distribution shapes unlikely matching a perfect normal distribution. This underlines the nature of the approximation.
We suggest to make this aspect the subject of a future study in order to learn if e.g. the asymmetric shape in \cref{fig:allseeds} is a statistic effect (combination of multiple binomial distributions including correlations as the training data in folds overlaps) or systematic with respect to the loss landscape optimization.

\begin{figure}[h]
    \centering
    \begin{subfigure}[b]{\textwidth}
        \includegraphics[width=\textwidth]{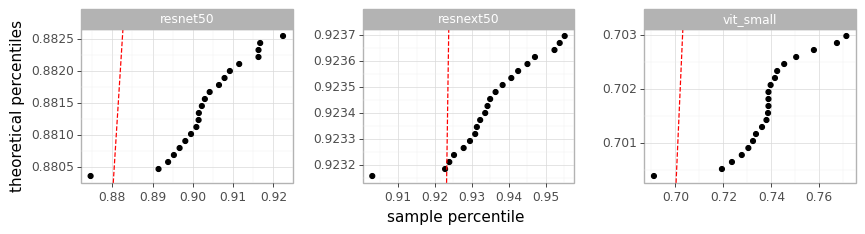}
        \caption{One seed}
        \label{fig:seed42qq}
    \end{subfigure}
    \newline
    \begin{subfigure}[b]{\textwidth}
        \includegraphics[width=\textwidth]{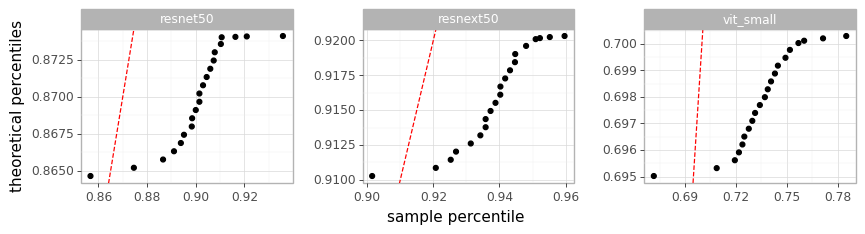}
        \caption{Six different seeds}
        \label{fig:allseedsqq}
    \end{subfigure}
    \caption{Quantile-Quantile plot of accuracy estimates for 10-class image classification for three different ML architectures across 20 folds and 6 seeds. \Cref{fig:seed42qq} shows results for one seed only (15 bins due to low statistics). \Cref{fig:allseedsqq} shows results across all 6 seeds (25 bins). The red dashed line marks the perfect correlation between a theoretical Gaussian distribution at the same $\mu, \sigma$ as obtained from the measured accuracy ensembles. }

    \label{fig:acc_qq}
\end{figure}

\Cref{fig:acc_qq} supports the notion expressed earlier, that neither ensembles at the presented size expose strong similarity to a Gaussian distribution. The quantiles obtained from the samples are apart the ideal correlation with the expected theoretical percentiles given a Gaussian distribution at the same mean $\mu$ and variance $\sigma$.

\subsubsection{Comparison of Sample Statistics and Approximated Confidence Intervals}
\label{sssec:confcompare}

We consider uncertainties introduced by seeds as a placeholder for more effects discussed by \citet{hooker}. We highlight at this point that such a procedure is impractical for realistic scenarios as it would require an insurmountable amount of computing resources, i.e. $360$ times the runtime. However, this provides the unique opportunity to compare a (realistic) single seed case versus a (unrealistic but desirable) multi-seed analysis.

We repeat the analysis alongside \cref{fig:accuracies} based on the estimates of all available folds and the obtained confidence intervals in order to motivate a judgment, whether approximate metric uncertainties can aid the reviewing process and how close they come to a representative estimate.

\begin{figure}[h]
    \centering
    \includegraphics[width=.8\textwidth]{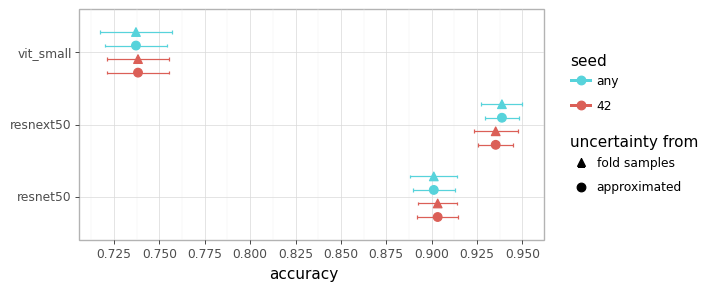}
    \caption{Comparison of \emph{fold sample} based uncertainty with \emph{approximated} uncertainty using \cref{eq:accconf}. Each estimate was obtained for one seed ($42$) or any seed available (total 6 seeds). The uncertainty plotted for seed $42$ was obtained using the approximation in \cref{eq:accconf}. The uncertainty plotted for all seeds was obtained using the sample standard deviation.}
    \label{fig:compare_noseed_approx}
\end{figure}

\Cref{fig:compare_noseed_approx} aspires to compare the uncertainty estimates obtained using from two different methods. One approach obtains uncertainties from fold samples using the sample standard deviation of accuracy measurements. The second approach uses approximation described in \cref{eq:accconf}.
In the single seed case (which is the norm in practice), uncertainty approximations in \cref{fig:compare_noseed_approx} are inline with fold sample estimates. This observation reproduces findings in previous reports \citep{Bouthillier}. This supports our conclusion that such approximations can offer a time and compute efficient method to obtain uncertainty estimates.

In the case of measurements entering from any seed, the uncertainty approximation from \cref{eq:accconf} appears to underestimate consistently the accuracy error estimate from fold samples. This is not unexpected. From our point of view, this demonstrates existing limitations of the method. We attribute this shortcoming to a violation of the underlying approximation assumptions of \cref{eq:accconf}. In this multi-seed multi-fold scenario, the sample of accuracy measurements contain another source of uncertainty in addition to the one described by \cref{eq:accconf} - for example stochastic effects introduced from several differently seeded training runs.

\subsection{Workflow and provenance of results}
\label{ssec:smk_fold10}

The design of this analysis using an automated snakemake \citep{smk} workflow bears several advantages. This technology allow execution of training and analysis on GPU or CPU and local or distributed hardware alike. It also offers the full provenance of a produced figure with respect to the training data.

\begin{sidewaysfigure}[ht]
    \centering
    \includegraphics[width=\textwidth]{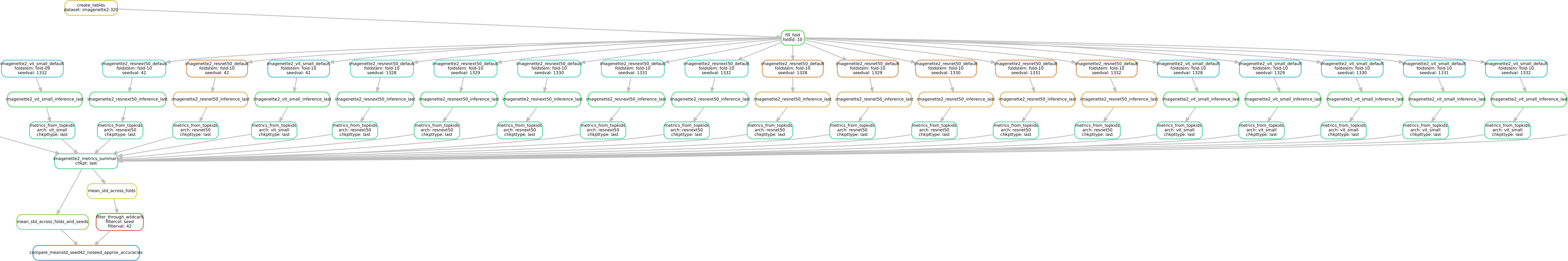}
        \caption{Portion of directed acyclic graph of tasks that lead to the creation of \Cref{fig:compare_noseed_approx}. For visibility, only tasks dependent on fold $10$ are shown.}

    \label{fig:smk_fold10}
\end{sidewaysfigure}

We explicitly highlight this functionality here as we believe that such infrastructure can support the review of ML workflows. \Cref{fig:smk_fold10} depicts the directed acyclic graph that lead to the creation of \Cref{fig:compare_noseed_approx} starting from the \texttt{imagenette2-320} dataset. We believe that jupyter notebooks are very helpful for practical needs during exploration.

They however often fall short when used as a platform for workflows or when sharing such workflows. Given tooling and visualization techniques, having automated workflows as a standard in the authoring process can greatly enhance the transparency of studies and therefor make reviewing easier if not possible at all.

All code to reproduce experiments is made available \citep{ourrepo}.

\end{document}